%% file: neurips_2023.tex
\documentclass{article}

\usepackage[final,nonatbib]{neurips_2023}

\usepackage[utf8]{inputenc} % allow utf-8 input
\usepackage[T1]{fontenc}    % use 8-bit T1 fonts
\usepackage{hyperref}       % hyperlinks
\usepackage{url}            % simple URL typesetting
\usepackage{booktabs}       % professional-quality tables
\usepackage{amsfonts}       % blackboard math symbols
\usepackage{nicefrac}       % compact symbols for 1/2, etc.
\usepackage{microtype}      % microtypography
\usepackage{xcolor}         % colors
\usepackage{graphicx}
\usepackage{multirow}
\usepackage{float}
\usepackage{authblk}
\usepackage{blindtext}

\title{NoPose-NeuS: Jointly Optimizing Camera Poses with Neural Implicit Surfaces for Multi-view Reconstruction}

\author[1]{Mohamed Shawky Sabae}
\author[1]{Hoda Anis Baraka}
\author[1,2]{Mayada Mansour Hadhoud}
\affil[1]{Faculty of Engineering, Cairo University}
\affil[2]{University of Science and Technology Zewail City}
\affil[ ]{\textit {\{mohamedshawky911, hoda.baraka, mayada.hadhoud\}@eng.cu.edu.eg}}

\begin{document}

\maketitle

\begin{abstract}
Learning neural implicit surfaces from volume rendering has become popular for multi-view reconstruction. Neural surface reconstruction approaches can recover complex 3D geometry that are difficult for classical Multi-view Stereo (MVS) approaches, such as non-Lambertian surfaces and thin structures. However, one key assumption for these methods is knowing accurate camera parameters for the input multi-view images, which are not always available. In this paper, we present NoPose-NeuS, a neural implicit surface reconstruction method that extends NeuS to jointly optimize camera poses with the geometry and color networks. We encode the camera poses as a multi-layer perceptron (MLP) and introduce two additional losses, which are multi-view feature consistency and rendered depth losses, to constrain the learned geometry for better estimated camera poses and scene surfaces. Extensive experiments on the DTU dataset show that the proposed method can estimate relatively accurate camera poses, while maintaining a high surface reconstruction quality with $0.89$ mean Chamfer distance.
\end{abstract}

\section{Introduction}
\input{core/intro}

\section{Related Work}
\input{core/related}

\section{Proposed Method}
\input{core/method}

\section{Experimental Results}
\input{core/results}

\section{Conclusion}
\input{core/conclusion}

\medskip

%Sets the bibliography style to UNSRT and imports the 
%bibliography file "samples.bib".
\bibliographystyle{unsrt}
\bibliography{neurips_2023}

%%%%%%%%%%%%%%%%%%%%%%%%%%%%%%%%%%%%%%%%%%%%%%%%%%%%%%%%%%%%

\end{document}

%% file: core/intro.tex
3D reconstruction from multi-view images is a fundamental problem in computer vision and computer graphics. Traditionally, 3D reconstruction pipelines, such as COLMAP \cite{schoenberger2016sfm} \cite{schoenberger2016mvs}, contain multiple steps to generate multi-view depth maps and fuse them into a dense point cloud representation, which is then used to recover scene surfaces. These methods rely on correspondence matching between RGB images, which causes artifacts and missing regions due to matching errors. Following the recent advances in NeRF \cite{mildenhall2020nerf}, new approaches \cite{oechsle2021unisurf} \cite{yariv2021volume} \cite{wang2023neus} have emerged to learn neural implicit surfaces using volume rendering. Generally, these works aim to jointly optimize implicit geometry and color networks guided by photometric loss from differentiable volume rendering. The geometry network is often encoded as a multi-layer perceptron (MLP) representing a signed distance function (SDF).

A common assumption in classical and neural rendering based techniques is the existence of accurate camera parameters for the input multi-view images. Actual camera parameters are not always easy to obtain in real situations, such as in casually captured images. Consequently, Structure from Motion (SfM) \cite{schoenberger2016sfm} methods are used to estimate camera parameters and sparse 3D points before performing multi-view geometry optimization. Recent NeRF-based methods \cite{wang2022nerf} \cite{lin2021barf} \cite{jeong2021selfcalibrating} \cite{bian2023nopenerf} propose joint optimization techniques of camera parameters with radiance fields.

\begin{figure}[t]
\begin{center}
\begin{tabular}{c c c c c}
\includegraphics[width=.2\linewidth]{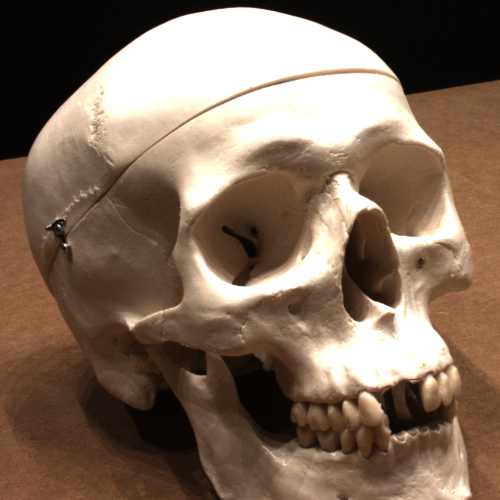} & \includegraphics[width=.2\linewidth]{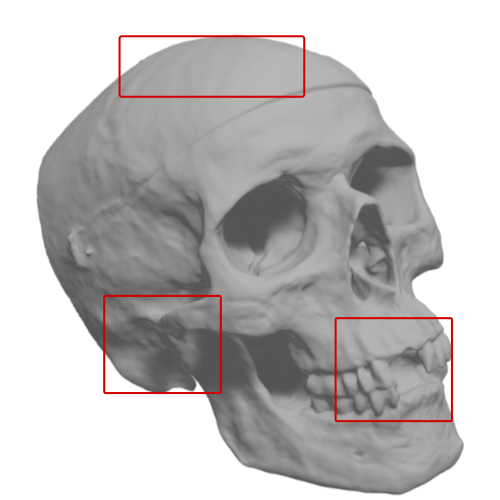} & \includegraphics[width=.2\linewidth]{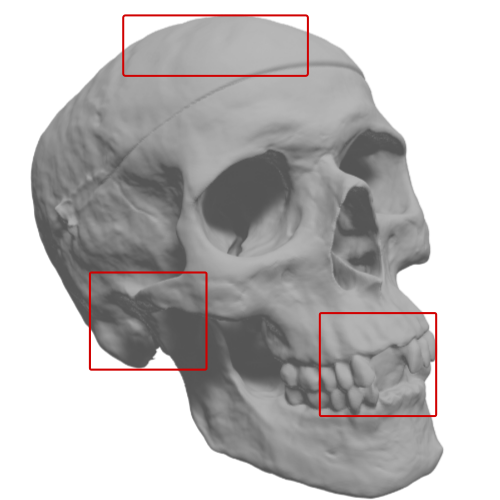} &
\includegraphics[width=.2\linewidth]{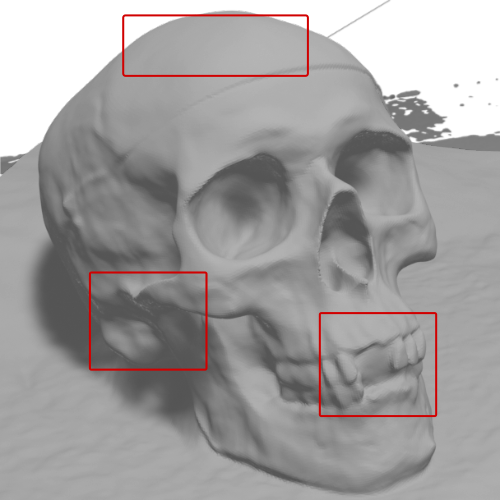} & \includegraphics[width=.2\linewidth]{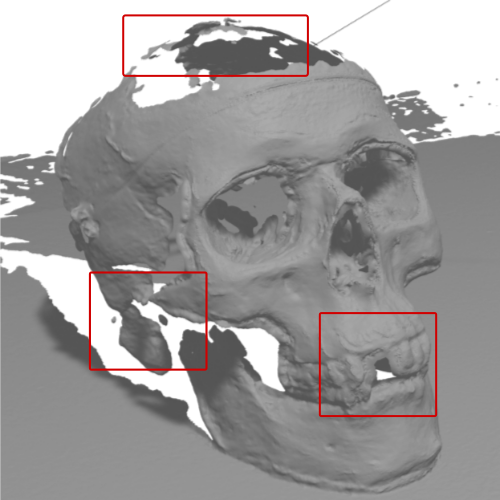} \\
Reference Image & \textbf{NoPose-NeuS} {\tiny (ours)} & NeuS \cite{wang2023neus} & MonoSDF \cite{yu2022monosdf} & COLMAP \cite{schoenberger2016mvs} 
\end{tabular} \\\
\caption{An example of reconstructed surfaces from the DTU dataset \cite{aanaes2016large}, showing the reconstruction quality of our proposed method compared to other methods.}
\label{fig:showcase}
\end{center}
\end{figure}

In this work, we enable camera pose optimization in SDF-based surface reconstruction methods, particularly NeuS \cite{wang2023neus}. Following NeRFtrinsic Four \cite{schieber2023nerftrinsic}, we use an MLP to predict camera poses from Gaussian Fourier features of camera indices. We impose two additional constraints, which are multi-view feature consistency, inspired by MVSDF \cite{zhang2021learning} and D-NeuS \cite{chen2022recovering}, and rendered depth loss. These additional losses constrain the learned geometry to jointly optimize camera poses along with geometry and color networks. Results on the DTU dataset \cite{aanaes2016large} show that the proposed method estimates camera poses with high relative accuracy, while outperforming classical MVS methods in terms of reconstruction quality, and offering comparable quality to other SDF-based surface reconstruction methods that depend on accurate input camera parameters, as shown in Figure \ref{fig:showcase}.

\setlength\tabcolsep{1pt}

The main contributions of this paper are summarized as follows:

\begin{itemize}
    \item We propose a joint optimization method of camera poses with geometry and color networks of NeuS \cite{wang2023neus} constrained by additional multi-view feature consistency and rendered depth losses, in order to maintain high quality geometry compared to other SDF-based surface reconstruction methods relying on input camera poses.
    \item We evaluate our proposed method both qualitatively and quantitatively on the DTU dataset \cite{aanaes2016large} and show high surface reconstruction quality with relatively accurate camera poses compared with other baselines.
\end{itemize}

%% file: core/related.tex
\subsection{Classical Multi-view Stereo}
Traditional Multi-view Stereo (MVS) methods \cite{schoenberger2016mvs} \cite{5226635} \cite{Galliani2016GipumaMP} aim to recover a global 3D dense representation of the scene from a set of posed overlapping images. These methods estimate the pixel-wise depth map of each input image using pairwise matching of RGB image patches, and then fuse the depth maps into a dense point cloud. As a post-processing step, surfaces are recovered from the dense point cloud using methods like Screened Poisson surface reconstruction \cite{10.1145/2487228.2487237}. As these methods require camera parameters for the input RGB images, Structure from Motion (SfM) \cite{schoenberger2016sfm} is first applied to recover the camera parameters and the sparse point cloud representation of the scene. The reconstruction quality of the classical MVS methods is heavily affected by the quality of the correspondence matching, which is usually difficult for regions without rich textures.

Recently, deep learning-based MVS methods \cite{yao2018mvsnet} \cite{wang2020patchmatchnet} \cite{zhang2020visibilityaware} showed better performance in estimating depth maps and dense representations. Furthermore, transformer-based approaches \cite{ding2021transmvsnet} \cite{cao2022mvsformer} can capture long-range context information across images, further enhancing the quality of the reconstruction. Similar to classical MVS methods, learning-based methods rely on input camera parameters and struggle with low-texture regions. In this work, we relax the assumption of having input camera extrinsics (poses) guided by multi-view feature consistency, inspired by MVS approaches, to constrain volume rendering for surface reconstruction.

\subsection{Neural Implicit Surface Representation and Reconstruction}
Neural implicit representations have gained increasing attention recently, because of the ability to learn continuous and highly-complex functions using simple neural networks. Implicit 3D scene representation can encode 3D scenes with high spatial resolution into a multi-layer perceptron (MLP) with few layers. Consequently, this representation is successfully applied to multi-view 3D reconstruction. The related works in this area can be roughly categorized into surface rendering based methods and volume rendering based methods. Surface rendering based methods \cite{niemeyer2020differentiable} \cite{yariv2020multiview} render the pixel color as the color of the intersection between the pixel ray and the scene geometry, which is vulnerable to self-occlusions and sudden changes in depth. On the other hand, volume rendering based methods \cite{mildenhall2020nerf} render the pixel color using alpha-composition of sampled point colors along the pixel ray. These methods are more robust to complex geometry changes, however the learned geometry is often noisy, due to the lack of constraints on the geometry level sets.

To mitigate such issues, methods, such as UNISURF \cite{oechsle2021unisurf}, VolSDF \cite{yariv2021volume} and NeuS \cite{wang2023neus}, learn implicit geometry functions, represented as occupancy values or signed distance functions (SDF), using volume rendering. SDF-based methods \cite{yariv2021volume} \cite{wang2023neus} are generally better, because the surface can be extracted as the zero-level set of the SDF using the Marching Cubes algorithm \cite{10.1145/37402.37422}, which produces more accurate geometry. Similar to classical MVS, these methods assume knowing accurate camera parameters for the input images.

Following NeRF \cite{mildenhall2020nerf}, methods, such as NeRFmm \cite{wang2022nerf}, BARF \cite{lin2021barf}, SC-NeRF \cite{jeong2021selfcalibrating} and NoPe-NeRF \cite{bian2023nopenerf}, propose techniques to jointly optimize camera parameters with radiance fields to relax the assumption of knowing accurate camera parameters by imposing additional losses to the optimization process. In this work, we enable camera pose optimization in SDF-based surface reconstruction methods, particularly NeuS \cite{wang2023neus}, by combining better camera parameterization and additional losses to constrain the learned geometry.

%% file: core/method.tex
\begin{figure}[t]
    \begin{center}
        \fbox{\includegraphics[scale=0.15]{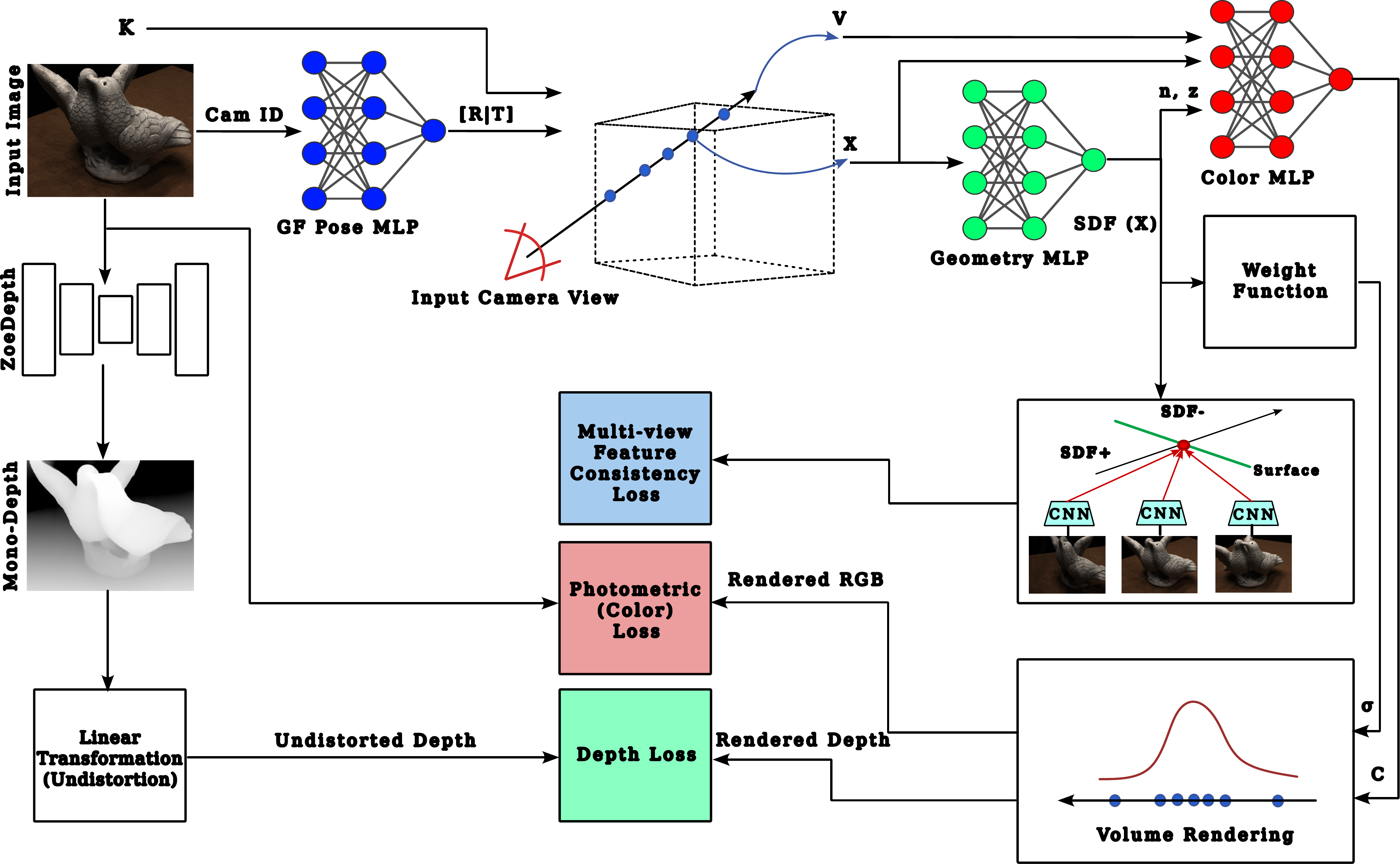}}
        \caption{Overview of the proposed method. We aim to jointly learn camera poses along with scene geometry and color functions following the formulation of NeuS \cite{wang2023neus}. We use an MLP to predict the camera poses from Gaussian Fourier features of camera indices \cite{schieber2023nerftrinsic}. Furthermore, we impose two additional constraints: 1) multi-view feature consistency from MVSDF \cite{zhang2021learning} and D-NeuS \cite{chen2022recovering}, 2) monocular depth supervision using predicted depth maps from ZoeDepth \cite{bhat2023zoedepth}.}
        \label{fig:nopose-neus-arch}
    \end{center}
\end{figure}

Given a set of RGB images of a scene, the goal is to reconstruct its surface represented by the zero-level set of an implicit neural signed distance field (SDF) without knowing the camera poses or alternatively given rough initial camera poses. The overview of the proposed method is illustrated in Figure \ref{fig:nopose-neus-arch}. We allow camera poses to be jointly optimized along with implicit neural networks. In this section, we first review the usage of volume rendering to learn SDF-based implicit neural surfaces. Then, we explain our parameterization of the camera poses and the two additional losses for improving the estimated camera poses and the overall surface quality.

\subsection{SDF-based Surface Reconstruction using Volume Rendering}

Using volume rendering to learn SDF-based implicit surfaces combines the advantages of surface rendering based and volume rendering based methods, where the scene space is constrained by a signed distance field. The surface $S$ is then represented as the zero-set level of an implicit SDF field $S = \{ x \in \mathbb{R}^3 | f(x) = 0 \}$, where $f$ is a function $f : \mathbb{R}^3 \rightarrow \mathbb{R}$ that maps a spatial position $x \in \mathbb{R}^3$ to its signed distance to the object surface. This mapping function is implicitly encoded using a multi-layer perceptron (MLP), which encodes the scene geometry. In addition to the MLP of scene geometry (SDF), another MLP is used to encode the color function $g : \mathbb{R}^3 \times \mathbb{S}^2 \times \mathbb{S}^2 \times \mathbb{R}^{N_f} \rightarrow \mathbb{R}^3$ that maps a spatial point $x \in \mathbb{R}^3$, its viewing direction $v \in \mathbb{S}^2$, its normal surface $n \in \mathbb{S}^2$ calculated by differentiating the SDF function $\nabla f(x)$, and a feature vector $v \in \mathbb{R}^{N_f}$ generated by the SDF network to the point color $c \in \mathbb{R}^3$. In order to train both networks, 3D points are sampled along a ray emitted through an image pixel as follows:

\begin{equation}
    x(t) = o + tv | t \geq 0
    \label{eq:ray}
\end{equation}

where $o$ is the camera origin, $v$ is the unit vector of the ray direction, and $t$ is the distance between the 3D point $x$ and the camera origin $o$. Then, sampled colors are weighted and accumulated along the ray using volume rendering in order to get the pixel color to compare with the ground-truth:

\begin{equation}
    C(o, v) = \int_{0}^{+\infty} w(t) g(x(t), v, n, z) dt
    \label{eq:render}
\end{equation}

where $C(o, v)$ is the output color of the corresponding pixel, $g(x(t), v, n, z)$ is the color of the 3D point $x(t)$ given the viewing direction $v$, and $w(t)$ is the weight of the 3D point $x(t)$. NeuS \cite{wang2023neus} introduced a weight function that is both unbiased and occlusion-aware:

\begin{equation}
    w(t) = exp\left(-\int_{0}^{t} \rho(u)du\right) \rho(t)
\end{equation}

where $\rho(t)$ is called an opaque density function, which is the counterpart of the density function $\sigma(t)$ in the standard volume rendering. Equation \ref{eq:render} can be approximated using discretization; refer to NeuS \cite{wang2023neus} for more details on this. The rendered pixel colors are then compared with the ground-truth input image pixel colors for supervising the networks training.

\subsection{Camera Parameterization}

The camera parameters $\Pi = K[R | t]$ include camera intrinsics $K_i$, which transform the points from the camera $i$ coordinates into the image $i$ coordinates, and extrinsics (poses) $T_i = [R_i | t_i]$, which transform the world coordinates into camera $i$ coordinates. We assume that the camera intrinsics $K$ are known, as they are usually included in the image metadata. We only consider optimizing the camera poses $[R | t]$ in this work, where $R \in SO(3)$, $t \in \mathbb{R}^3$ and $[R | t] \in SE(3)$.

Following NeRFtrinsic Four \cite{schieber2023nerftrinsic}, the index of each camera is mapped to a higher-dimensional space using Gaussian Fourier feature mapping from Tancik et al. \cite{tancik2020fourier}.

\begin{equation}
    \gamma(v) = [cos(2 \pi Bv), sin(2 \pi Bv)]^T, B \in \mathbb{R}^{m \times d}
\end{equation}

where $v$ is the low-dimensional input and $B$ is a matrix for the Gaussian mapping, whose values are sampled from $N(0, \sigma^2)$ and frequency parameter $m$.

These features are then passed through an MLP with GELU activation functions to predict the pose for each camera, which contains translation vector $t \in \mathbb{R}^3$ and rotation vector in axis-angle representation $r \in so(3)$ that is used to construct the rotation matrix $SO(3)$. This formulation gives the camera poses more degrees of freedom to learn than directly optimizing rotation and translation vectors, which enables joint optimization of camera, geometry, and color networks.

\subsection{Multi-view Consistency}

Generally, using multi-view consistency constraints is common in 3D reconstruction methods. This becomes more important when optimizing camera poses, as the multi-view consistency constraints ensure correct relative poses between cameras. Photo-consistency approaches use photometric distance across RGB images, which is typically used in classical MVS methods \cite{schoenberger2016mvs}. Meanwhile, feature consistency approaches compare pixels in feature maps of different views.

We use feature consistency on surface points to constrain camera and surface optimization. Similar to D-NeuS \cite{chen2022recovering}, we utilize the SDF values of the sampled 3D points to extract the surface points. This is done using linear interpolation to find the zero-crossing of the SDF values between the last positive and the first consecutive negative SDF values.

Once the surface points are obtained, features of this point are compared across multiple views, similar to MVSDF \cite{zhang2021learning}. Features are extracted by applying a convolutional neural network (CNN), pretrained for supervised MVS \cite{zhang2020visibilityaware}, on the RGB images. The final multi-view feature consistency loss is formulated as follows:

\begin{equation}
    L_{feature} = \frac{1}{N_c N_s} \sum_{i=1}^{N_s} |F_0(p_0) - F_i(K_i(R_i x^{'} + t_i))|
\end{equation}

where $N_c$ is the number of channels in the feature maps, $N_s$ is the number of neighboring source views, $F$ is the extracted feature map for a specific view, $p_0$ is the pixel through which the ray is cast in the reference view, $x_{'}$ is the interpolated surface point, and $K_i(R_i x^{'} + t_i)$ is the surface point projected on the source view $i$ using its camera parameters ${K_i, R_i, t_i}$.

It is clear that the multi-view feature consistency loss imposes direct constraint on the predicted camera poses ${R_i, t_i}$ to enforce correct relative pose between the reference and source cameras.

\subsection{Depth Supervision}

In order to improve the quality of the reconstructed surface and keep the estimated camera translation within reasonable limits, we apply monocular depth loss against ground-truth depth maps. Ground-truth depth maps $\bar{D}$ are predicted by applying a pretrained monocular depth predictor on the input RGB images. We choose ZoeDepth \cite{bhat2023zoedepth} for monocular depth prediction, which offers state-of-the-art monocular depth quality. However, the predicted depth maps from the RGB images are usually not multi-view consistent.

Consequently, we use monocular depth undistortion technique from Nope-NeRF \cite{bian2023nopenerf}, which considers learning scale and shift parameters $\{(\alpha_i, \beta_i) | i = 0...N-1\}$ for each view. These parameters are used to linearly transform monocular depth maps to recover multi-view consistent depth maps $\bar{D}^*$ to be used for depth supervision:

\begin{equation}
    \bar{D}_i^* = \alpha_i \bar{D} + \beta_i
\end{equation}

The scale $\alpha_i$ and shift $\beta_i$ parameters are jointly optimized along with camera poses, geometry and color MLPs. Furthermore, the predicted depth maps $\hat{D}$ are obtained using volume rendering as follows:

\begin{equation}
    \hat{D} = \int_{0}^{+\infty} w(t) t dt
\end{equation}

Similar to Equation \ref{eq:render}, $w(t)$ is the weight of the 3D point $x(t)$ and $t$ is the distance between the distance between the 3D point $x(t)$ and the camera origin $o$ from Equation \ref{eq:ray}.

The rendered depth maps $\hat{D}$ are then compared with the undistorted ground-truth depth maps $\bar{D}_i^*$ using L1 loss on $N_r$ rays (pixels) in the minibatch:

\begin{equation}
    L_{depth} = \frac{1}{N_r} \sum_{j}^{N_r} || \hat{D}_j - \bar{D}_j^* ||_1
\end{equation}

\subsection{Overall Training Loss}

The overall loss to jointly optimize camera poses, depth undistortion parameters, geometry network and color network is formulated as follows:

\begin{equation}
    L = L_{rgb} + \lambda_1 L_{eikonal} + \lambda_2 L_{mask} + \lambda_3 L_{feature} + \lambda_4 L_{depth}
    \label{eq:total_loss}
\end{equation}

The color loss $L_{rgb}$ is defined as L1 photometric loss between the rendered RGB $\hat{C}$ and the input RGB images $C$ on $N_r$ rays (pixels) in the minibatch:

\begin{equation}
    L_{rgb} = \frac{1}{N_r} \sum_{j}^{N_r} || \hat{C}_j - C_j ||_1
\end{equation}

Moreover, Eikonal loss $L_{eikonal}$ is applied on the sampled points to regularize the gradients of the SDF field predicted by the geometry network $f$:

\begin{equation}
    L_{eikonal} = \frac{1}{N_r N_p} \sum_{j, k}^{N_r, N_p} (|| \nabla f(x_{j,k}) ||_2 - 1)^2 
\end{equation}

where $N_r$ is the number of rays (pixels) in the minibatch, and $N_p$ is the number of sampled 3D points per ray. Also, we use an optional mask loss $L_{mask}$, which is defined as binary cross entropy loss against ground-truth mask, as described in the original NeuS paper \cite{wang2023neus}.

Finally, we apply coarse-to-fine optimization from BARF \cite{lin2021barf}, which adds increasingly higher frequencies to the positional encoding of both 3D position and viewing direction during training. This is proven to reduce the likelihood of converging to a local minimum for camera pose optimization.

%% file: core/results.tex
\subsection{Dataset}
To evaluate our method against other baselines, we use the DTU dataset \cite{aanaes2016large}, which is widely-used for evaluating 3D reconstruction methods with a challenging variety of geometry, materials and appearance. We choose the same $15$ scenes as those used in IDR \cite{yariv2020multiview} and NeuS \cite{wang2023neus}. Each scene contains $48$ or $64$ images of a resolution of $1200$ $\times$ $1600$. Ground truth camera parameters are also provided to evaluate our estimated camera poses using relative pose error (RPE) between pairs of image views. Furthermore, reference point clouds for all scenes are provided in the dataset for quantitative evaluation using the Chamfer distance provided by the official dataset evaluation protocol.

\subsection{Experimental Setup}

\textbf{Baselines.} We compare the quality of our reconstructed geometry to the widely-used classical MVS pipeline COLMAP \cite{schoenberger2016sfm} \cite{schoenberger2016mvs}, as well as other SDF-based surface reconstruction methods: NeuS \cite{wang2023neus} and MonoSDF \cite{yu2022monosdf}, which offer state-of-the-art results on the DTU dataset. For MonoSDF, we use the MLP representation trained on all input views for fair comparison. Moreover, we quantitatively compare our estimated camera poses with those estimated by COLMAP using the relative pose error (RPE) between pairs of image views.

\textbf{Implementation Details.} Similar to NeuS \cite{wang2023neus}, the geometry network contains $8$ hidden layers, each of size $256$ and a skip connection from the input to the output of the $4$th layer. The color (radiance) network contains $4$ hidden layers, each of size $256$. We follow the same hierarchical sampling strategy for volume rendering from NeuS. Positional encoding is applied to position $x$ and viewing direction $v$, with frequencies of $6$ and $4$, respectively. However, we use a coarse-to-fine scheduling strategy, similar to BARF \cite{lin2021barf}, to smoothly mask the frequencies of positional encoding on an interval $[0.1, 0.5]$. Moreover, the camera pose network is modeled as an MLP with $3$ layers with a hidden size of $64$ and GELU activation functions, similar to NeRFtrinsic Four \cite{schieber2023nerftrinsic}. The input frequency parameter $m$ is set to $128$, which results in an embedding size of $256$. We allow the estimation of actual camera poses or relative transformation from an initial pose. By default, all cameras are zero initialized at the center of a unit sphere, following NeuS initialization of the SDF network. For multi-view feature consistency loss, we use $N_s = 2$ where each reference view is compared to two source views using $N_c = 32$ feature channels. We train our method for $300$k iterations and $512$ sampled rays per batch for $21$ hours on a single Nvidia RTX $2080$ti GPU. We use the ADAM optimizer \cite{kingma2017adam} with a learning rate of $5e-4$. Also, we set $\lambda_1$, $\lambda_2$, $\lambda_3$ and $\lambda_4$ to $0.1$, $0.1$, $0.5$ and $0.01$, respectively. After optimization, we use the Marching Cubes algorithm \cite{10.1145/37402.37422} to extract mesh from the learned SDF field using bounding boxes defined by estimated camera poses with volume size of $512^3$ voxels.

\subsection{Results}

We conducted our comparisons with baselines both quantitatively and qualitatively. In Table \ref{table:chamfer_distance}, we report the Chamfer distances on the selected scenes from the DTU dataset \cite{aanaes2016large}. The results show that our method offers comparable results to NeuS \cite{wang2023neus} and MonoSDF \cite{yu2022monosdf}, which are SDF-based surface reconstruction methods that rely on input camera parameters. Meanwhile, our method outperforms the classical MVS pipeline COLMAP \cite{schoenberger2016mvs} in most cases. Furthermore, we show the relative pose errors in Table \ref{table:relative_pose_error} against COLMAP, which is the only method in our baselines that optimizes camera poses. The results show that our method maintains accurate relative poses between different views as good as the classical MVS pipelines. This is mainly due to the imposed constraints by the multi-view feature consistency and depth losses. Note that the estimated camera poses are sensitive to initialization, as discussed in \ref{sec:discussion}.

\setlength\tabcolsep{1.5pt}
\renewcommand{\arraystretch}{1.2}
\begin{table}[ht]
\begin{center}
\begin{tabular}{c c c c c c c c c c c c c c c c c c c}
\hline
Scan & 24 & 37 & 40 & 55 & 63 & 65 & 69 & 83 & 97 & 105 & 106 & 110 & 114 & 118 & 122 & \textbf{Mean} \\
\hline
COLMAP \cite{schoenberger2016mvs} & \textbf{0.81} & 2.05 & 0.73 & 1.22 & 1.79 & 1.58 & 1.02 & 3.05 & 1.40 & 2.05 & 1.00 & 1.32 & 0.49 & 0.78 & 1.17 & 1.36 \\
NeuS \cite{wang2023neus} & 1.00 & \textbf{1.37} & 0.93 & \textbf{0.43} & 1.10 & 0.65 & \textbf{0.57} & 1.48 & \textbf{1.09} & 0.83 & 0.52 & 1.20 & \textbf{0.35} & \textbf{0.49} & \textbf{0.54} & \textbf{0.84} \\
MonoSDF \cite{yu2022monosdf} & 0.83 & 1.61 & \textbf{0.65} & 0.47 & \textbf{0.92} & 0.87 & 0.87 & \textbf{1.30} & 1.25 & \textbf{0.68} & 0.65 & \textbf{0.96} & 0.41 & 0.62 & 0.58 & \textbf{0.84} \\
\hline
\textbf{NoPose-NeuS} & 0.91 & 1.51 & 0.95 & 0.44 & 1.01 & \textbf{0.63} & 0.79 & 1.53 & 1.22 & 0.88 & \textbf{0.51} & 1.35 & 0.39 & 0.55 & 0.67 & 0.89 \\
\hline
\end{tabular} \\\
\caption{Quantitative evaluation on the DTU dataset using the Chamfer distance (lower values are better). Results of the baselines are reported from the original papers, except for COLMAP results, which are taken from MonoSDF paper \cite{yu2022monosdf}. The best score for each scan is marked in \textbf{bold}.}
\label{table:chamfer_distance}
\end{center}
\end{table}

\setlength\tabcolsep{1.5pt}
\renewcommand{\arraystretch}{1}
\begin{table}[ht]
\begin{center}
\begin{tabular}{c c c c c c c c c c c c c c c c c c c}
\hline
\multicolumn{2}{c}{Scan} & 24 & 37 & 40 & 55 & 63 & 65 & 69 & 83 & 97 & 105 & 106 & 110 & 114 & 118 & 122 & \textbf{Mean} \\
\hline
\\
\multirow{2}{*}{\rotatebox[origin=c]{90}{RPE$_r$}} & COLMAP \cite{schoenberger2016mvs} & \textbf{0.52} & 0.93 & \textbf{0.62} & 0.71 & 0.72 & 0.65 & \textbf{0.61} & 0.89 & 0.66 & \textbf{0.71} & 0.55 & \textbf{0.64} & \textbf{0.57} & 0.57 & 0.63 & 0.67 \\
 & \textbf{\textbf{NoPose-NeuS}} & 0.55 & \textbf{0.89} & 0.69 & \textbf{0.54} & \textbf{0.66} & \textbf{0.54} & 0.62 & \textbf{0.79} & \textbf{0.57} & 0.75 & \textbf{0.48} & 0.66 & 0.60 & \textbf{0.51} & \textbf{0.58} & \textbf{0.63} \\\\
\hline
\\
\multirow{2}{*}{\rotatebox[origin=c]{90}{RPE$_t$}} & COLMAP \cite{schoenberger2016mvs} & \textbf{0.95} & \textbf{1.01} & 1.12 & 0.99 & 0.81 & 0.93 & \textbf{0.67} & 1.05 & \textbf{1.15} & 1.05 & 0.91 & 0.88 & \textbf{0.91} & 0.86 & 0.99 & 0.95 \\
 & \textbf{NoPose-NeuS} & 1.01 & 1.08 & \textbf{0.91} & \textbf{0.88} & \textbf{0.76} & \textbf{0.80} & 0.87 & \textbf{1.02} & 1.19 & \textbf{1.04} & \textbf{0.79} & \textbf{0.82} & 0.96 & \textbf{0.85} & \textbf{0.97} & \textbf{0.93} \\\\
\hline
\end{tabular} \\\
\caption{Quantitative evaluation of the estimated camera poses using relative pose error (lower values are better). We compare our estimated poses to those estimated by COLMAP in terms of relative rotation and translation errors. The rotation error (RPE$_r$) is reported in degrees, and the translation error (RPE$_t$) is scaled by 100. The best score for each scan is marked in \textbf{bold}.}
\label{table:relative_pose_error}
\end{center}
\end{table}

\begin{figure}[ht]
\begin{center}
\begin{tabular}{c c c c c c}
\rotatebox{90}{Scan24} & \includegraphics[width=.2\linewidth]{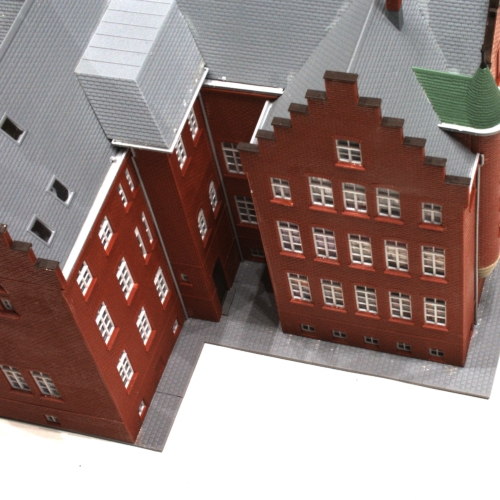} & \includegraphics[width=.2\linewidth]{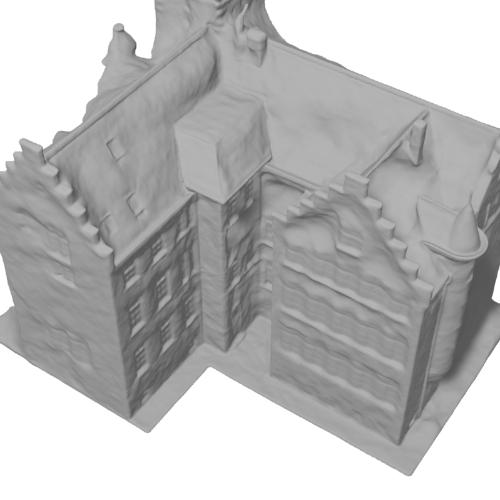} & \includegraphics[width=.2\linewidth]{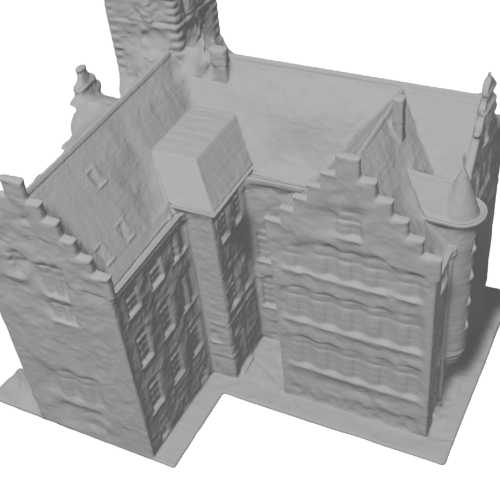} &
\includegraphics[width=.2\linewidth]{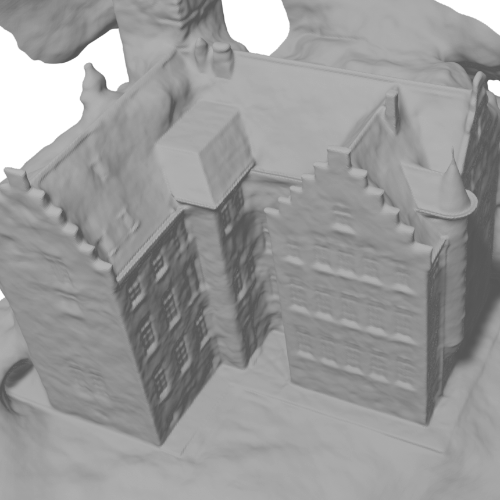} & 
\includegraphics[width=.2\linewidth]{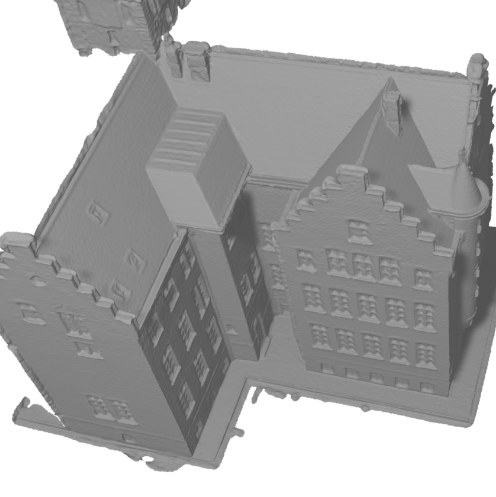} \\
\rotatebox{90}{Scan37} & \includegraphics[width=.2\linewidth]{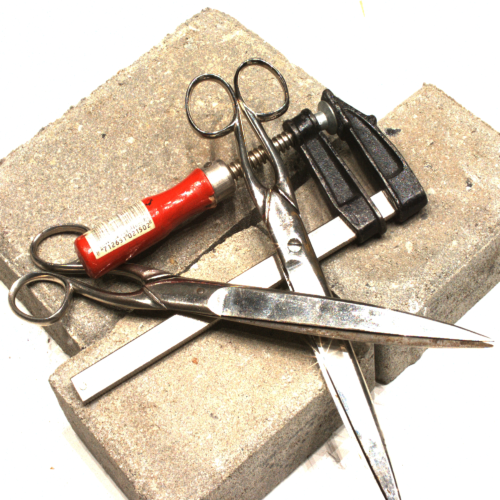} & \includegraphics[width=.2\linewidth]{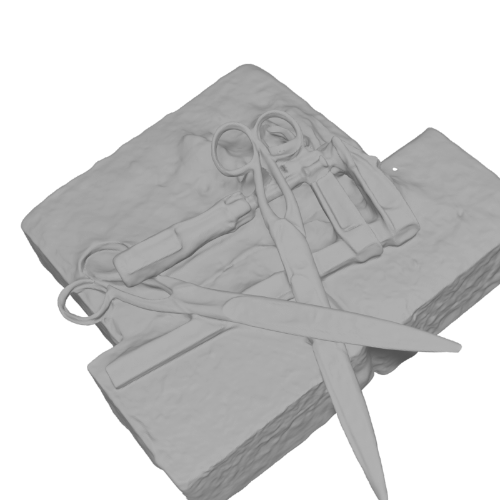} & \includegraphics[width=.2\linewidth]{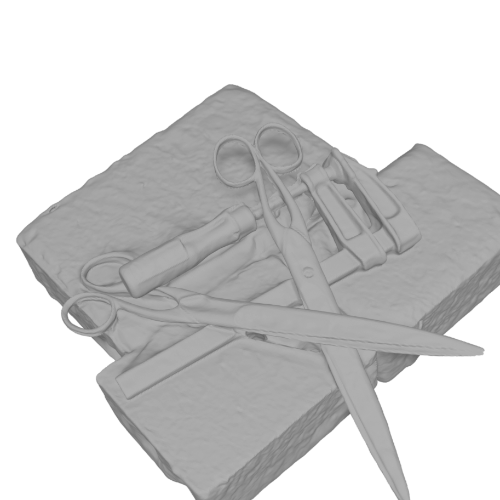} &
\includegraphics[width=.2\linewidth]{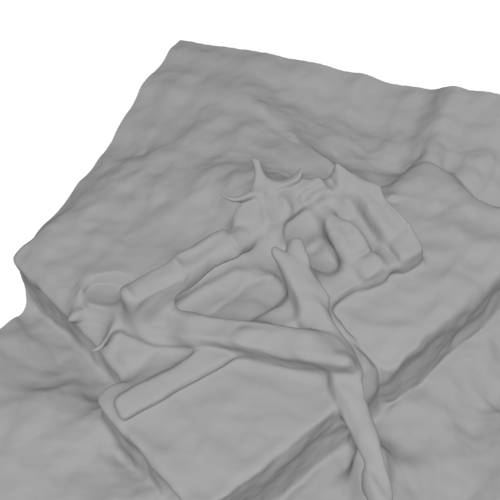} & 
\includegraphics[width=.2\linewidth]{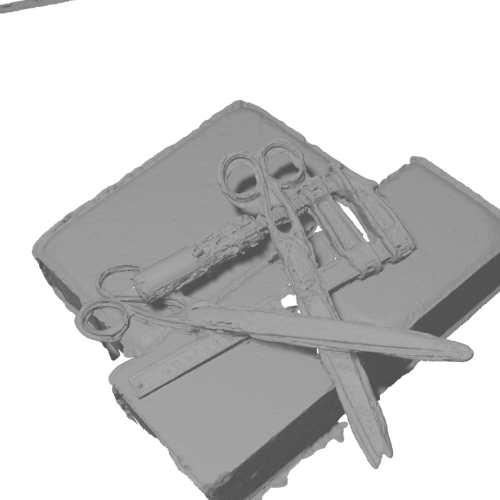} \\
\rotatebox{90}{Scan55} & \includegraphics[width=.2\linewidth]{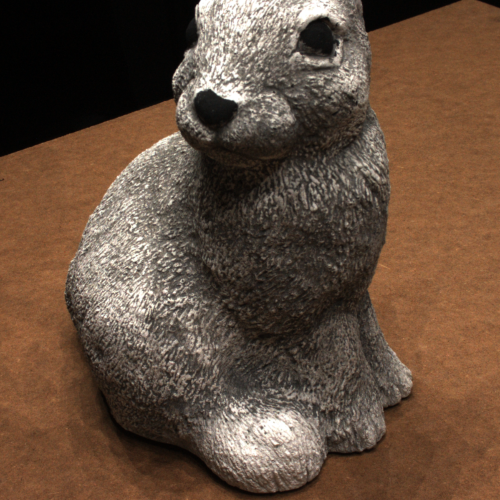} & \includegraphics[width=.2\linewidth]{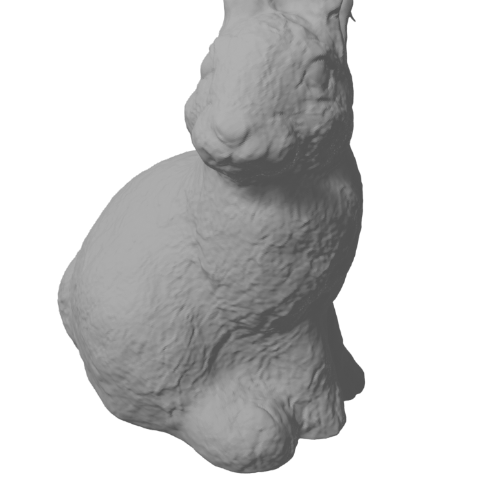} & \includegraphics[width=.2\linewidth]{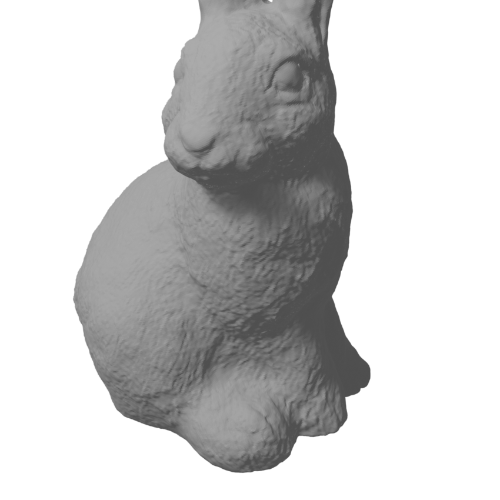} &
\includegraphics[width=.2\linewidth]{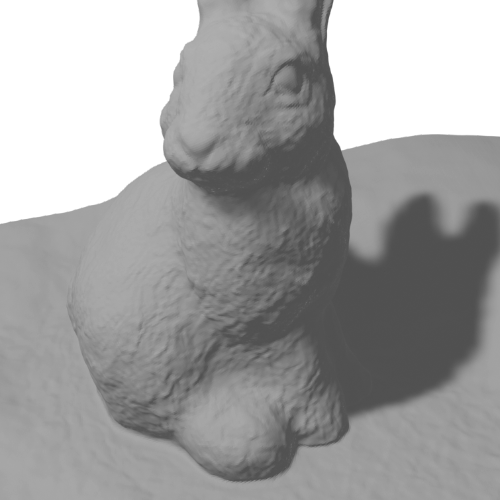} & 
\includegraphics[width=.2\linewidth]{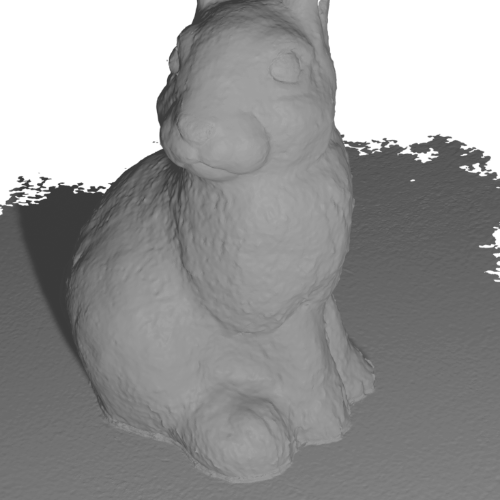} \\
\rotatebox{90}{Scan63} & \includegraphics[width=.2\linewidth]{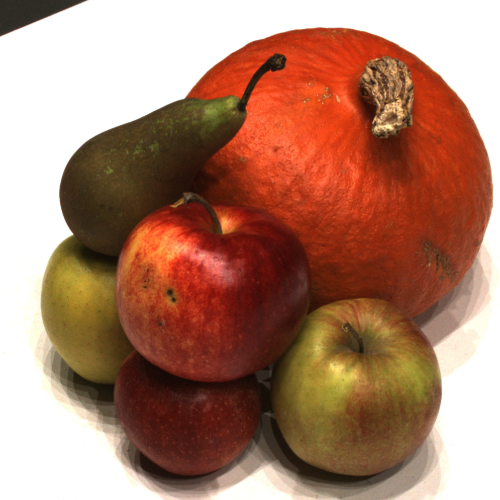} & \includegraphics[width=.2\linewidth]{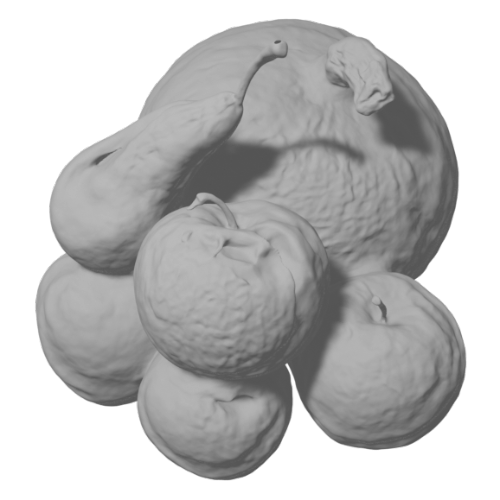} & \includegraphics[width=.2\linewidth]{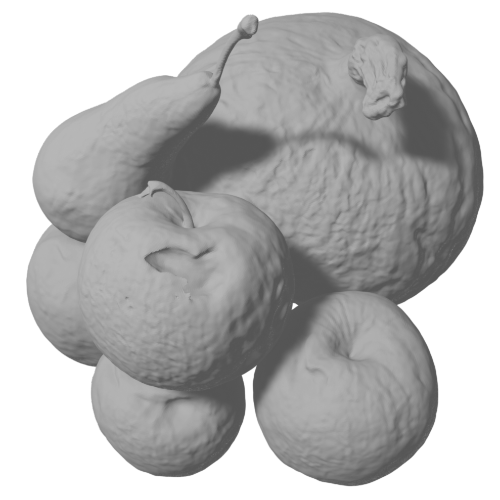} &
\includegraphics[width=.2\linewidth]{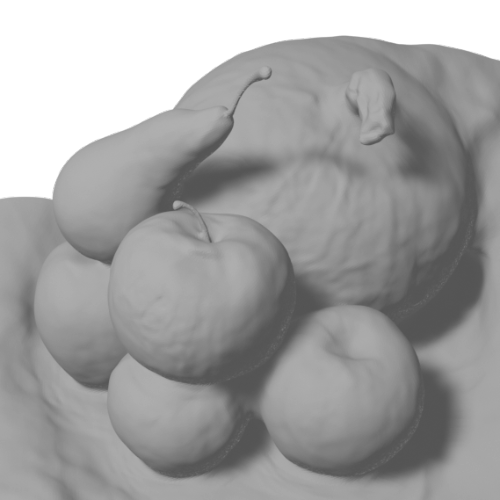} & 
\includegraphics[width=.2\linewidth]{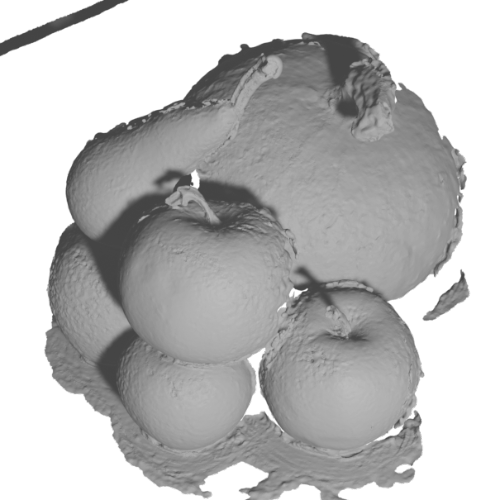} \\
\rotatebox{90}{Scan65} & \includegraphics[width=.2\linewidth]{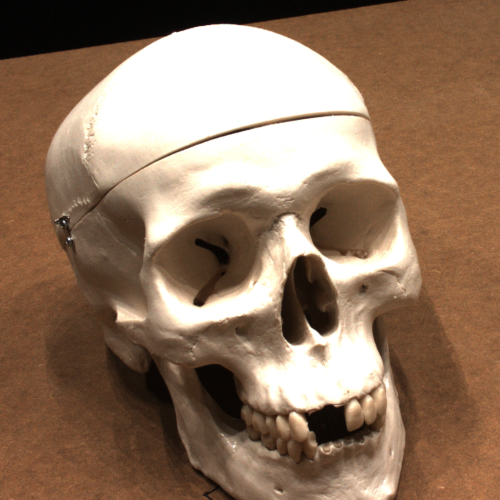} & \includegraphics[width=.2\linewidth]{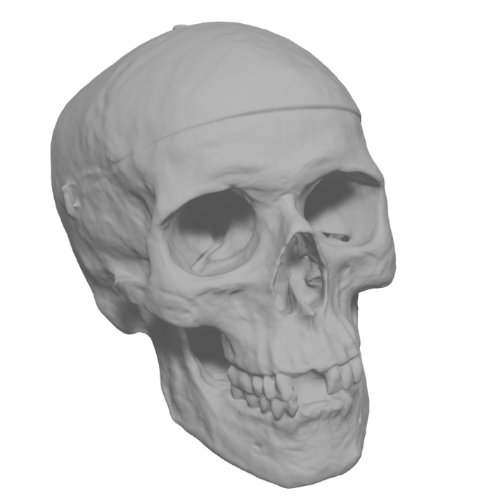} & \includegraphics[width=.2\linewidth]{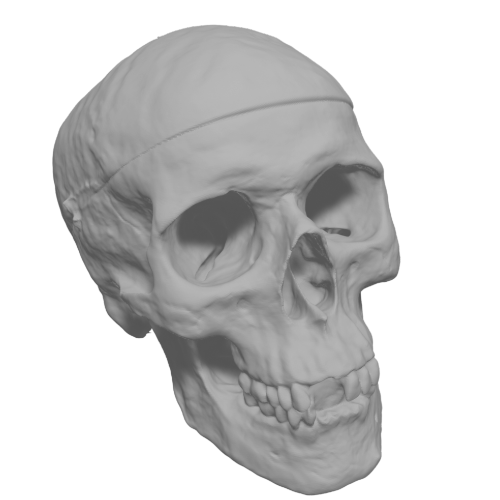} &
\includegraphics[width=.2\linewidth]{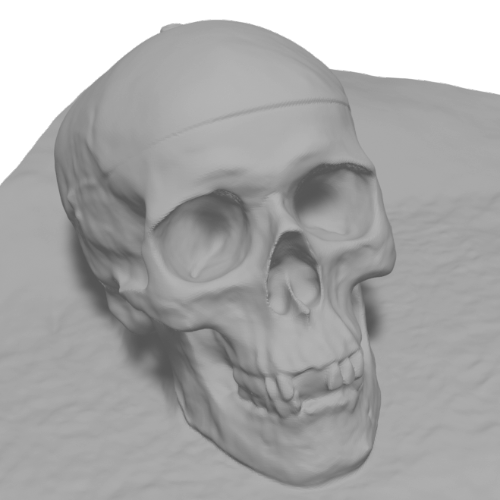} & 
\includegraphics[width=.2\linewidth]{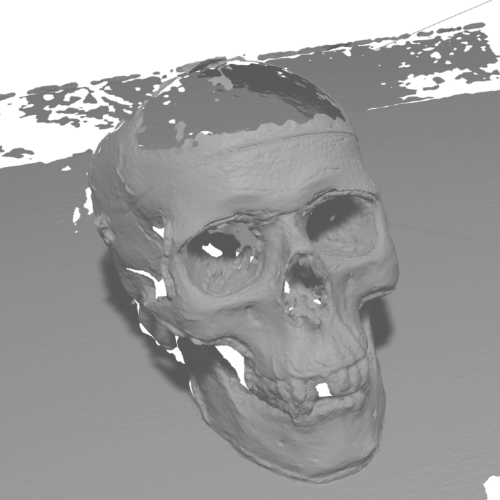} \\
\rotatebox{90}{Scan106} & \includegraphics[width=.2\linewidth]{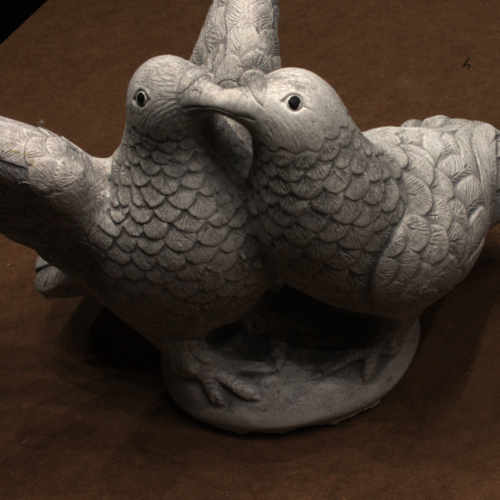} & \includegraphics[width=.2\linewidth]{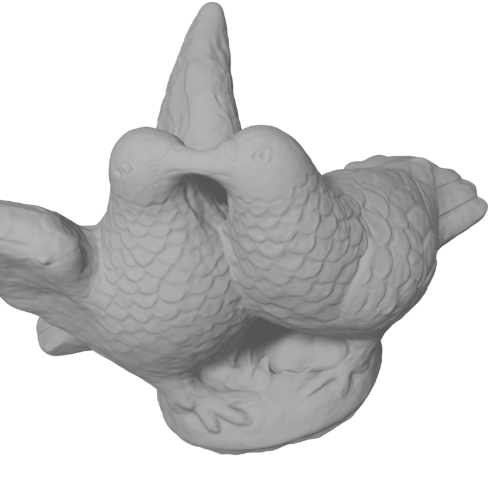} & \includegraphics[width=.2\linewidth]{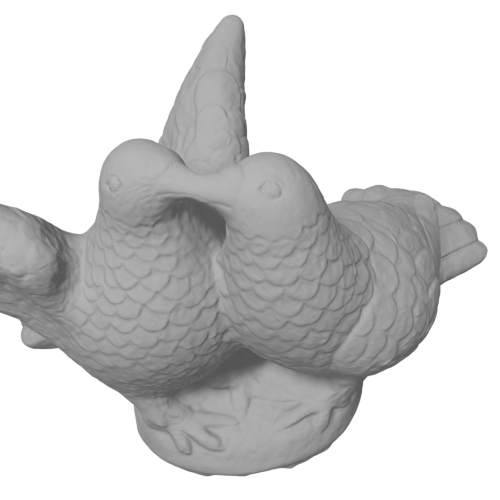} &
\includegraphics[width=.2\linewidth]{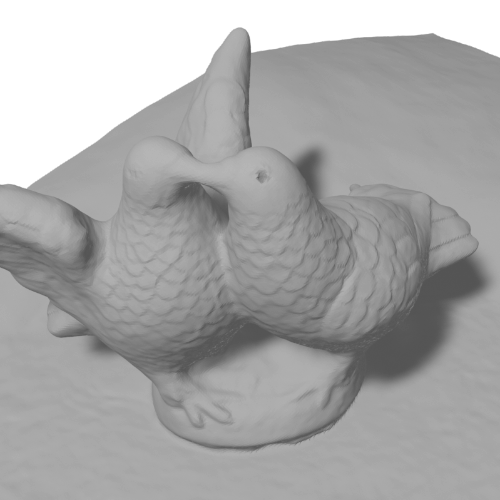} & 
\includegraphics[width=.2\linewidth]{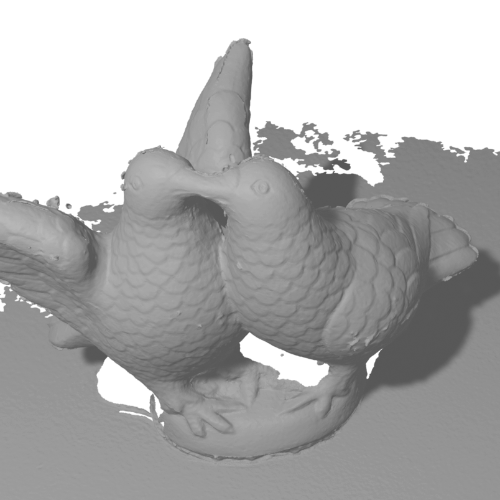} \\
& Reference Image & \textbf{NoPose-NeuS} {\tiny (ours)} & NeuS \cite{wang2023neus} & MonoSDF \cite{yu2022monosdf} & COLMAP \cite{schoenberger2016mvs} \\
\end{tabular} \\\
\caption{Qualitative evaluation of the surface reconstruction of our method against COLMAP \cite{schoenberger2016mvs}, NeuS \cite{wang2023neus} and MonoSDF \cite{yu2022monosdf}.}
\label{fig:visual_results}
\end{center}
\end{figure}

Moreover, Figure \ref{fig:visual_results} shows our qualitative results against the baselines. Our method is able to recover highly-accurate geometry, while jointly optimizing camera poses. COLMAP results suffer from noisy and discontinuous surfaces (scans 37, 65 and 106). Meanwhile, MonoSDF (MLP) can reconstruct smooth surfaces with high accuracy (scan 63), however it struggles with thin structures (scan 37). Our method reconstructs continuous surfaces with a high level of quality, which outperforms COLMAP in all cases. Moreover, our method can handle complex geometry (scans 24, 37, 65 and 106), offering comparable geometry quality to NeuS and MonoSDF (MLP), which rely on accurate input camera parameters. However, flat surfaces are better in NeuS in some cases (scan 24 and 37). Also, MonoSDF (MLP) is better for smooth surfaces in other cases (scan 63).

\setlength\tabcolsep{1pt}

\subsection{Discussion}
\label{sec:discussion}

\begin{figure}[ht]
\begin{center}
\begin{tabular}{c | c c | c c}
\includegraphics[width=.2\linewidth]{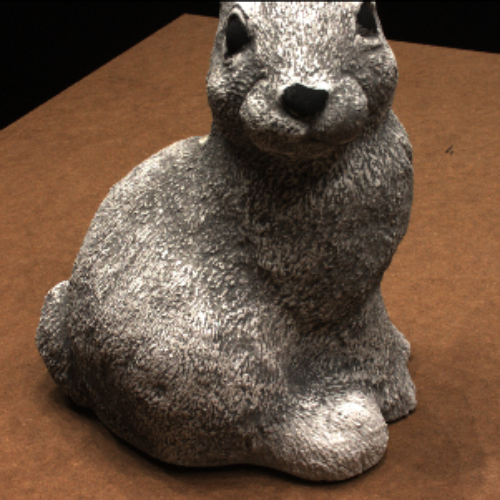} & \includegraphics[width=.2\linewidth]{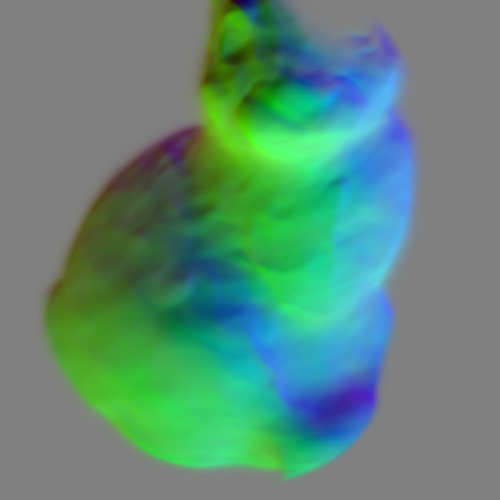} & \includegraphics[width=.2\linewidth]{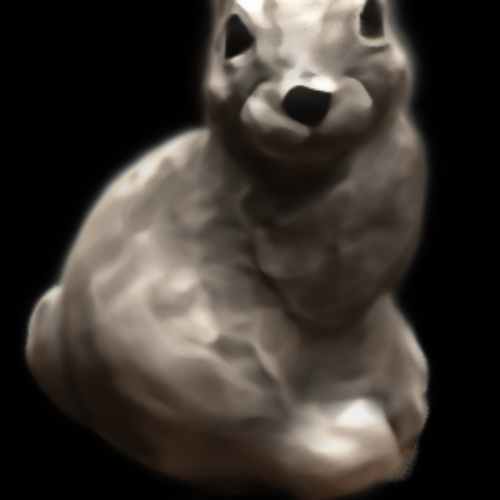} &
\includegraphics[width=.2\linewidth]{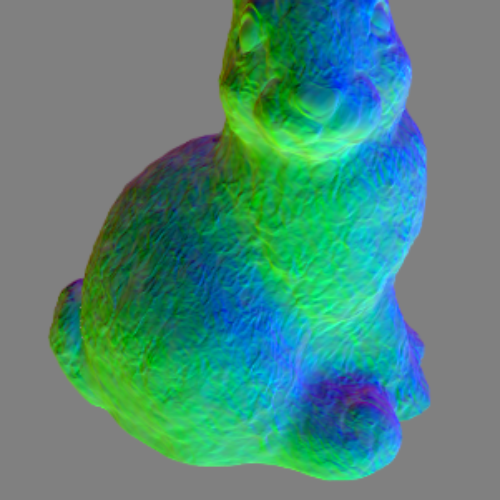} & \includegraphics[width=.2\linewidth]{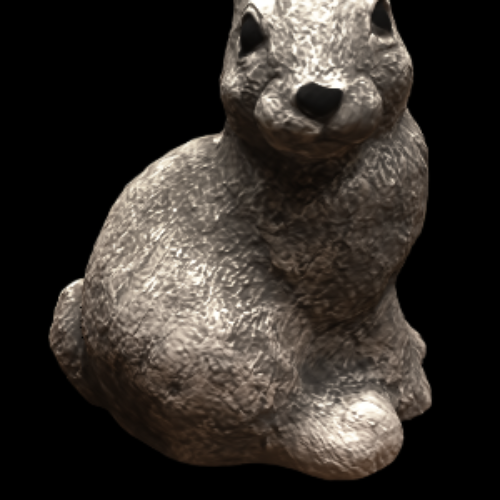} \\
Reference Image & \multicolumn{2}{c|}{a) Random Initial Poses} & \multicolumn{2}{c}{b) Centered Initial Poses} \\
\end{tabular} \\\
\caption{Qualitative results showing the effect of camera pose initialization on the reconstruction quality. We show rendered surface normal maps and RGB images of two models: a) initialized with random noisy camera poses, b) initialized with zero (centered) camera poses.}
\label{fig:pose_analysis}
\end{center}
\end{figure}

\textbf{Camera Pose Initialization.} Good camera pose initialization is essential for our method to achieve high-quality reconstruction. We performed an analysis to show the effect of camera initialization. As illustrated in Figure \ref{fig:pose_analysis}, initializing camera poses (translation vector $t \in \mathbb{R}^3$ and rotation vector $r \in so(3)$) with random values that do not follow any structure can result in losing fine details, as the optimization process cannot recover the correct relative poses between different views. However, it is better that the camera poses are zero initialized (center of a unit sphere), as the geometry network is initialized to produce an approximate SDF of a unit sphere. Note that near-optimal initialization of camera poses helps the model to converge faster to good results.

\setlength\tabcolsep{1pt}

\begin{figure}[ht]
\begin{center}
\begin{tabular}{c c c c c}
\includegraphics[width=.2\linewidth]{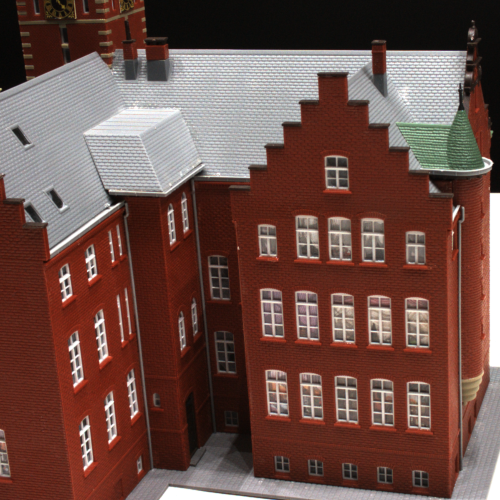} & \includegraphics[width=.2\linewidth]{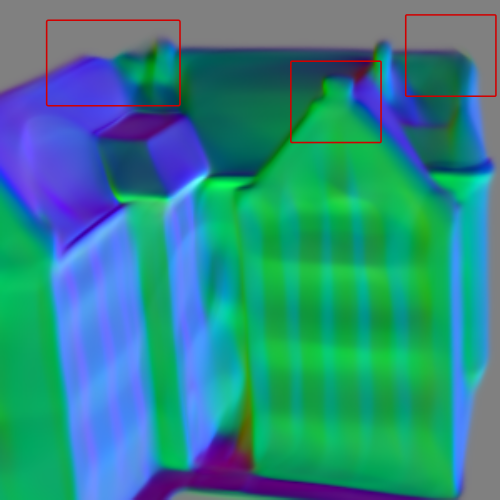} & \includegraphics[width=.2\linewidth]{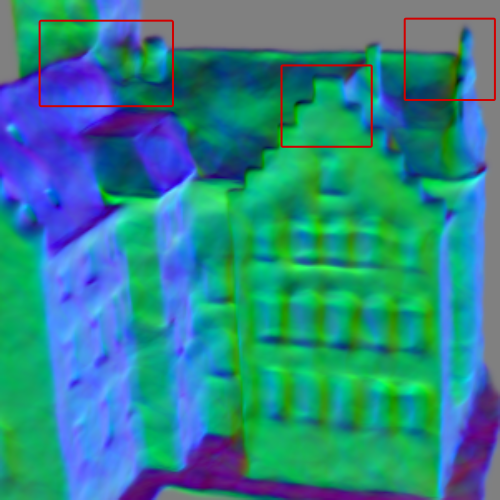} &
\includegraphics[width=.2\linewidth]{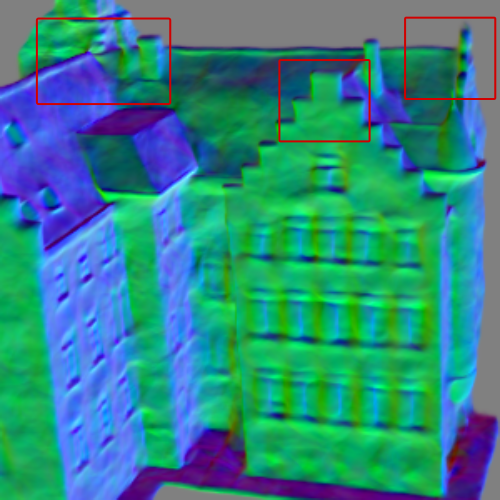} & \includegraphics[width=.2\linewidth]{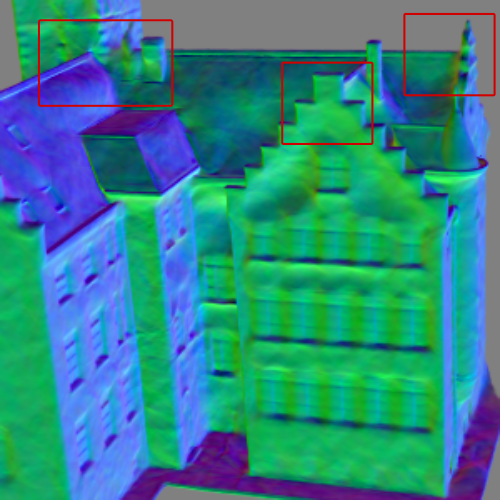} \\
Reference Image & W/o feature/depth & W/o feature & W/o depth & \textbf{Ours} 
\end{tabular} \\\
\caption{Qualitative results of the ablation study conducted on the DTU dataset. To further illustrate the importance of the two additional losses, we show the rendered normal maps in different cases.}
\label{fig:ablation}
\end{center}
\end{figure}

\textbf{Ablation Study.} We conducted an ablation study of the DTU dataset to evaluate the different components of our loss function. We started with the original losses from NeuS ($L_{rgb}$, $L_{eikonal}$ and $L_{mask}$), then progressively combined our additional losses. We show the rendered surface normal maps of DTU scan $24$ for different settings in Figure \ref{fig:ablation}. It is clear that using only the original NeuS losses results in over-smoothed geometry, while adding only feature consistency or depth loss results in noisy geometry. This is mainly due to high relative camera pose error. The final loss (described in Equation \ref{eq:total_loss}) offers the best reconstruction quality.

\setlength\tabcolsep{1pt}

%% file: core/conclusion.tex
We introduced NoPose-NeuS, a neural implicit surface reconstruction method that enables camera pose optimization in NeuS \cite{wang2023neus}. In our work, we encode the camera poses as an MLP, which is jointly optimized with the geometry and color networks. Furthermore, we impose two additional losses, which are multi-view feature consistency and rendered depth loss, to constrain the learned camera poses and 3D geometry. Our method can recover relatively accurate camera poses, while maintaining the quality of the surface reconstruction. The main limitation of our approach is the sensitivity to the camera initialization, as it assumes a bounded scene following NeuS formulation. Therefore, an interesting future work is to relax this assumption from the camera parameterization and the SDF network initialization. Moreover, we can further optimize the camera intrinsics for full camera calibration.